\documentclass[10pt,conference]{IEEEtran}

\usepackage{graphicx} 
\usepackage{subcaption}
\usepackage{amsmath}
\usepackage{mathtools}

\begin{document}

\title{Object Tracking with Correlation Filters using Selective Single Background Patch}

\author{\IEEEauthorblockN{Lasitha Mekkayil}
\IEEEauthorblockA{Dept. of Electronic and Communication Engineering\\
M. S. Ramaiah University of Applied Sciences\\
Bangalore, India \\
lasitha.ec.et@msruas.ac.in}\\ 
\and
\IEEEauthorblockN{Hariharan Ramasangu }
\IEEEauthorblockA{Dept. of Electronic and Communication Engineering\\
M.S. Ramaiah University of Applied Sciences\\
Bangalore, India \\
hariharan.ec.et@msruas.ac.in}\\
}

\maketitle

\begin{abstract}
Correlation filter plays a major role in improved tracking performance compared to existing trackers. The tracker uses the adaptive correlation response to predict the location of the target. Many varieties of correlation trackers were proposed recently with high accuracy and frame rates. The paper proposes a method to select a single background patch to have a better tracking performance. The paper also contributes a variant of correlation filter by modifying the filter with image restoration filters. The approach is validated using Object Tracking Benchmark sequences. 

\end{abstract}

\section{Introduction}
Tracking of an object through out the image plane forms object tracking. Robustness of the existing algorithms \cite{one}-\cite{ten} developed for tracking depends upon the ability to handle various challenges such as occlusion, illumination change, background clutter etc.  

Tracking algorithms can be broadly classified to generative \cite{one}-\cite{three} and discriminative algorithms \cite{four}-\cite{six}. Generative algorithms depends on a model to track the object and discriminative algorithms depends on the features to be extracted. The developed tracking algorithm includes algorithm based on correlation filters, oblique random forest, coarse to fine features, etc. 

In the current scenario, correlation filters forms a strong base for building object tracking algorithms due to its robust performance and high speed. Multiple tracking algorithms based on correlation filters were developed recently \cite{seven} - \cite{seventeen}. Though the correlation filter is a booming area among tracking algorithms, it possess various drawbacks. Some of the  drawback includes the circulant assumption, boundary suppression, and drift during tracking. A recently proposed context aware tracker uses multiple background patches for tracking \cite{eight}.

The paper proposes an algorithm for the selection of a single background patch with a restoration based correlation filter. The proposed approach discusses the tracking of a single object using correlation filter. The algorithm uses a new strategy for selection of background patches among the background pool. The selected background patches is utilised for locating the object of interest by excluding the other background patches from the background pool. The proposed algorithm is validated using precision and success score for Object Tracking Benchmark (OTB Dataset). From the validated results we can conclude that the proposed algorithm performs well compared to existing algorithms.

\section{Related Work}
Various Correlation Filter (CF) based trackers have developed recently due to its outstanding performance in tracking.  Bolme et al. \cite{seven} modeled the target appearance by an adaptive correlation filter which was optimized by minimizing the output sum of squared error (MOSSE). The convolution theorem can be used with correlation filters to accelerate tracking. Circulant structure with kernels tracker (CSK),
proposed by Henriques et al. \cite{twelve}, exploited the circular structure of adjacent subwindows
in an image for quickly learning a kernelized regularized least squares classifier of the target
appearance with dense sampling. Kernelized correlation filters (KCF) \cite{fifteen} was an extended
version of CSK by re-interpreting correlation tracking using the kernelized Ridge regression
with multi-channel features. Danelljan \cite{sixteen} introduced color attributes to improve tracking performance in colorful sequences and then proposed the DSST tracker \cite{fourteen} with accurate scale estimation by one separate filter. Zhang et al. \cite{eighteen} utilized the spatial-temporal context in the Bayesian framework to interpret correlation tracking. All of the discussed trackers attempt to exploit different characteristics with correlation filters for tracking, e.g. circular structure
\cite{twelve}, kernel trick \cite{fifteen}, color attributes \cite{sixteen}, effective feature representation (e.g. HOG) \cite{fourteen},
the consistency of object representation in scale space and context information \cite{eight}.


\section{Preliminaries}

The goal of correlaion filter is to learn a discriminative correlation filter that can be applied to the region of interest in consecutive frames to infer the location of the target (i.e. location of maximum filter response). The key contribution
leading to the popularity and success of CF trackers
is their sampling method. Due to computational constraints, it is common practice to randomly pick a limited number of negative samples around the target. The sophistication of the sampling strategy and the number of negative samples can have a significant impact on tracking performance. CF trackers allow for dense sampling around the target at very low computational cost. This is achieved by modeling all
possible translations of the target within a search window as circulant shifts and concatenating them to form the data matrix $P_0$. The circulant structure of this matrix facilitates a very efficient solution to the following ridge regression problem in the Fourier domain.

\begin{equation}
\min_w \left \| P_{0}w-y \right \|_{2}^{2}+ \lambda _{1}\left \| w \right \|_{2}^{2}
\end{equation} 

Here, the learned correlation filter is denoted by the vector w. The square matrix $P_0$ contains all circulant shifts of the vectorized image patch and the regression target y is a vectorized image of a 2D Gaussian.

A context aware correlation filter \cite{eight} discusses the importance of the surroundings and their impact on the tracked object while tracking. For example, if there is a lot of background clutter, background of the target is very important for successful tracking. The tracker selects $P_0 \epsilon R^n$ circulant patches of the target and $n$ background patches $P_i$ is selected. The selected background patches are called as hard negative patches. These hard negative patches forms distractors while tracking the target.  The algorithm learn a filter $w \epsilon R^n$ that has a high response for the target patch and close to zero response for background patches. The CA tracker implements the zero response of background patches by adding the background patches as a regularizer to the standard formulation. As a result, the target patch is regressed to y, while the background patches are regressed to zeros controlled by the parameter $\lambda_2$.

\begin{equation}
\min_w \left \| P_{0}w-y \right \|_{2}^{2}+ \lambda _{1}\left \| w \right \|_{2}^{2}+\lambda _{2}\sum_{i=1}^{n}\left \| P_iw \right \|_{2}^{2}
\end{equation}

\section{Proposed Algorithm}

The proposed algorithm is an extension of CA tracker. The algorithm discusses a way to select single background patch among the large set of $P_i$ background patches. The proposed algorithm also discusses the use of a restoration filter to obtain better tracking accuracy. The steps of the proposed algorithm is discussed below:

\begin{enumerate}
\item Identify the circulant shifted patches of the target  $P_0$ from $t^{th} frame$, calculate filter $w$ and gaussian shaped output $y$

\item Implement restoration of the filter by modifying the filter framework with a weiner filter or a constrained least square filter, $w_1$.

\item Calculate the background patches $P_i$ from frame $(t_1)^{th}$

\item For frame $t_1$, predict the target location using correlation filter $w1$ 

\item Identify the distance between selected background patches from step 3 and the location of the object to be tracked, identified from step 1 and 2.

\item Select the background patch with minimum distance measure and replace the set of background patches with a single background patch. 

\item Apply $w_1$  and the selected background patch identified from step 5 to predict the target final location for $t_1^{th} frame$

\item The final formulation of the filter also uses a patch selected from the target with a weight value, (0.25) to reduce the drift of the target. The background patches are regressed to zero by utilising a regulariser parameter $\lambda_1$,20. The values are selected based on multiple iterations and comparisons of the results generated. 

\end{enumerate}

\begin{table}[!b]
\renewcommand{\arraystretch}{1.3}
\caption{Performance comparison of multiple tracking dataset with respect to various trackers }
\label{table_example}
\centering
\begin{tabular}{| l | l | l | l | l | l | l | l |}
    \hline
    \textbf{Sequence} & \textbf{RCACF} & \textbf{CA} & \textbf{FCT} &  \textbf{VTD} & \textbf{IVT} & \textbf{MIL} & \textbf{Struck} \\ 
    \hline
    Tiger2 & 86.85 & 28.49 & 72 & 13 & 19 & 44 & 62 \\
    \hline
    Tiger1 & 100 & 90.14 & 52 & 78 & 8 & 34 & 87 \\
    \hline
   Animal & 97.18 & 100 & 92 & 96 & 4 & 87 & 96  \\
    \hline
   Jumping & 100 & 100 & 10 & 12 & 71 & 21 & 63  \\
    \hline
    Cliffbar & 98.48 & 87.88 & 99 & 53 & 47 & 71 & 8 \\
    \hline
    Woman & 22.39 & 17.91 & 11.6 & 14 & 17 & 15 & 67 \\
    \hline
    Surfer & 80.32 & 98.14 & 70 & 64 & 24 & 71 & 61 \\ 
    \hline
    Sylvester & 88.85 & 99.03 & 77 & 33 & 45 & 77 & 80 \\ 
    \hline
    Cardark & 100 & 100 & 36 & 25 & 54 & 48 & 18 \\
    \hline
    Faceocc1 & 100 & 100 & 43.4 & 80 & 74 & 60 & 82\\ 
    \hline
    Faceocc2 & 100 & 100 & 99 & 70 & 66 & 62 & 70 \\
    \hline
    Twinning & 94.68 & 100 & 98 & 54 & 42 & 60 & 77 \\
    \hline
    Box & 60.94 & 45.49 & 28.7 & 15 & 30 & 14 & 61 \\
     \hline
    \end{tabular}
   \end{table}

\section{Results and discussion}
The proposed tracker uses OTB 50 sequences for evaluation \cite{nineteen}.  The algorithm also uses 11 state-of-the-art-trackers for comparison of the performance analysis \cite{twenty}-\cite{twentysix}. Success plot and precision plot are the measures used to evaluate the tracker. The algorithm also explains the quantitative and qualitative analysis of the tracker.

\begin{figure} [!t]
\begin{center}  
\includegraphics[height=1.5in,width=3.5in,angle=0]{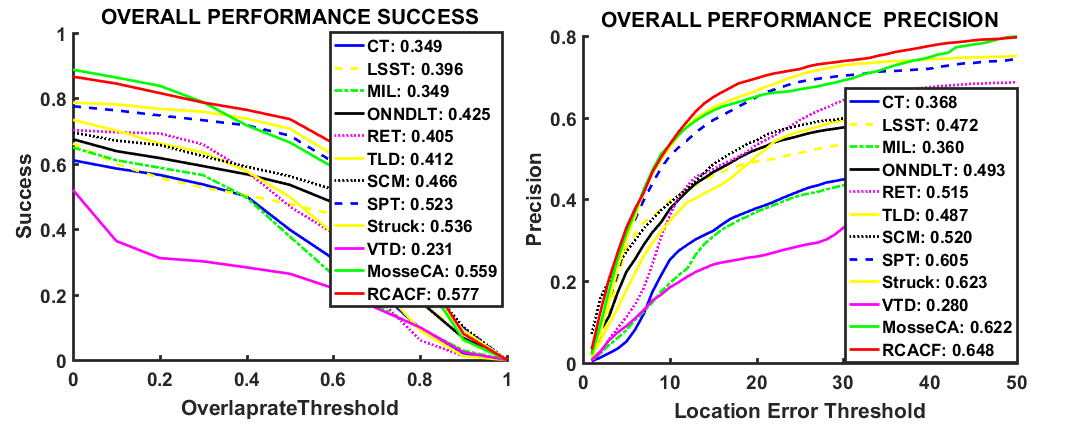}  
\caption{\small \sl Overall Success and Precision plot (Proposed RCACF algorithm is highlighted using red colour)
\label{fig:blk_fig}}  
\end{center}  
\end{figure}  

\begin{figure*}
\begin{subfigure}{.26\textwidth}
  \centering
  \includegraphics[height=1.5in,width=1.7in,angle=0]{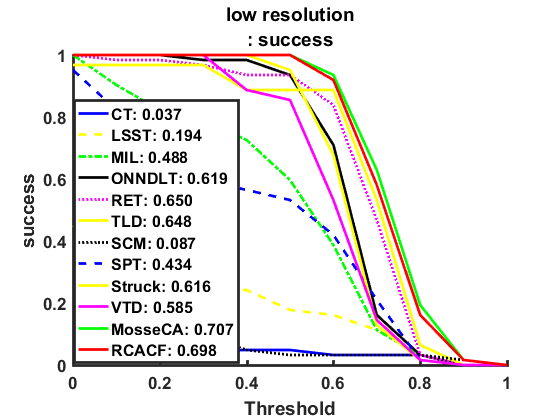}
  \label{fig:sfig1}
\end{subfigure}%
\begin{subfigure}{.2\textwidth}
  \centering
  \includegraphics[height=1.5in,width=1.7in,angle=0]{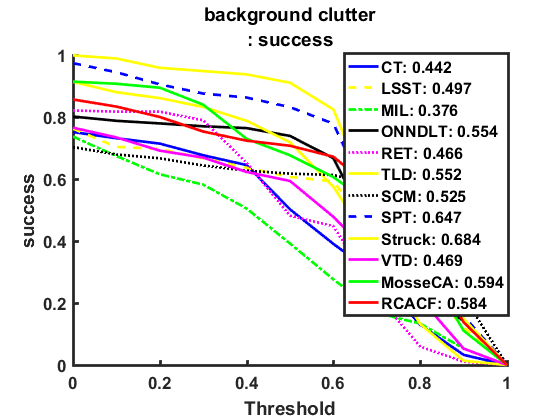}
  \caption{1b}
  \label{fig:sfig2}
\end{subfigure}
\begin{subfigure}{.2\textwidth}
  \centering
  \includegraphics[height=1.5in,width=1.7in,angle=0]{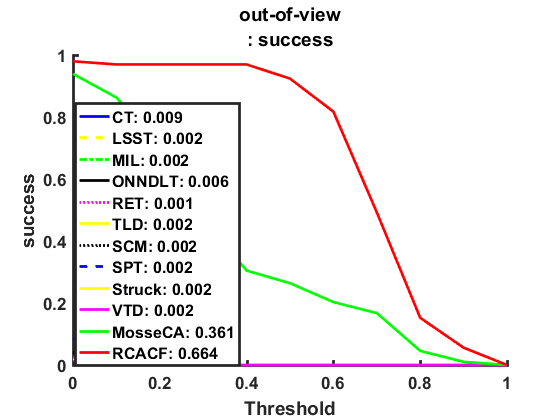}
  \label{fig:sfig2}
\end{subfigure}
\begin{subfigure}{.2\textwidth}
  \centering
  \includegraphics[height=1.5in,width=1.7in,angle=0]{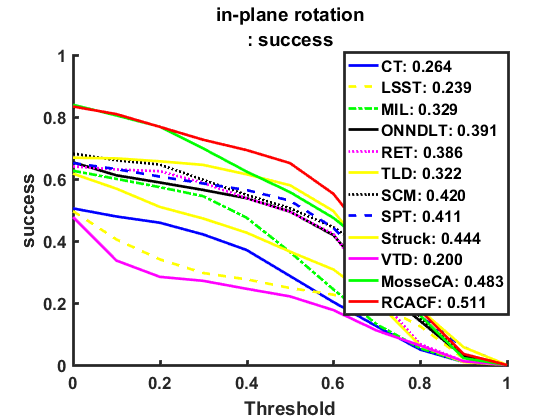}
  \label{fig:sfig2}
\end{subfigure}
\begin{subfigure}{.2\textwidth}
  \centering
  \includegraphics[height=1.5in,width=1.7in,angle=0]{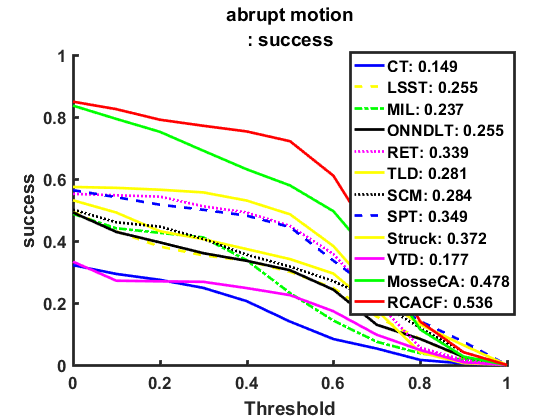}
  \label{fig:sfig2}
\end{subfigure}
\begin{subfigure}{.2\textwidth}
  \centering
  \includegraphics[height=1.5in,width=1.7in,angle=0]{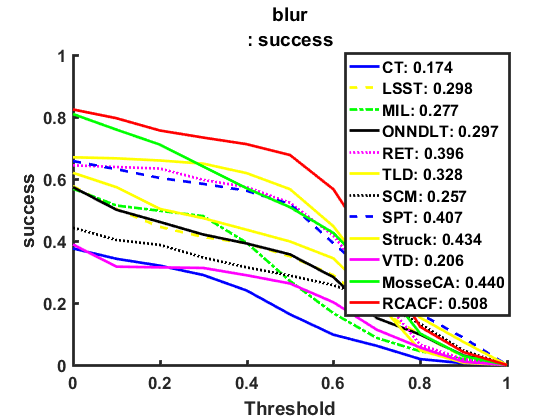}
  \label{fig:sfig2}
\end{subfigure}
\begin{subfigure}{.2\textwidth}
  \centering
  \includegraphics[height=1.5in,width=1.7in,angle=0]{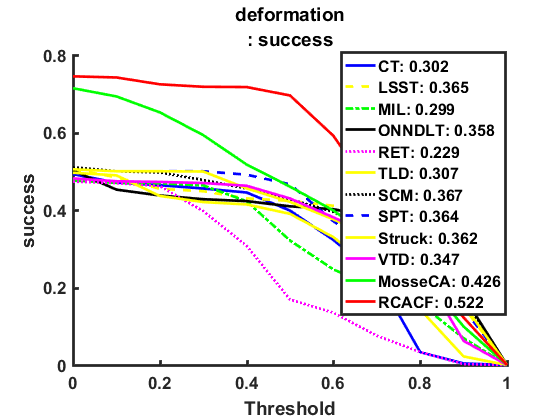}
  \label{fig:sfig2}
\end{subfigure}
\begin{subfigure}{.2\textwidth}
  \centering
  \includegraphics[height=1.5in,width=1.7in,angle=0]{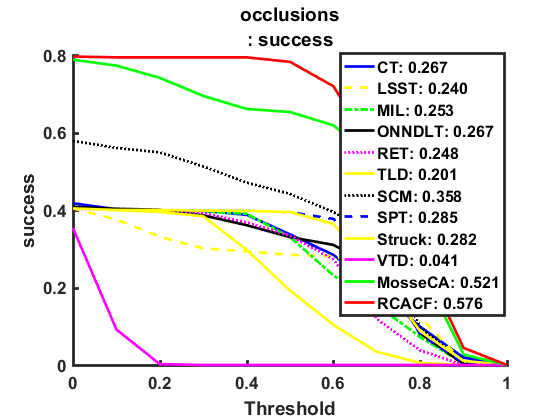}
  \label{fig:sfig2}
\end{subfigure}
\begin{subfigure}{.26\textwidth}
  \centering
  \includegraphics[height=1.5in,width=1.7in,angle=0]{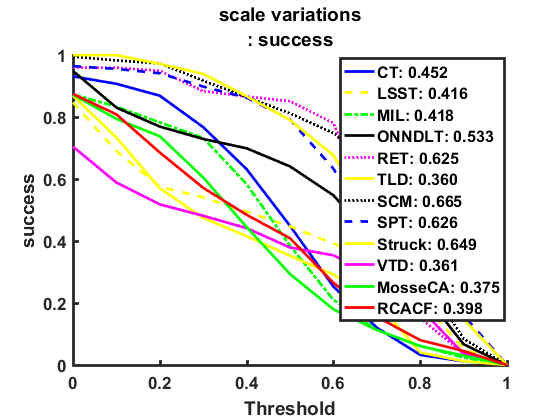}
  \label{fig:sfig2}
\end{subfigure}
\begin{subfigure}{.26\textwidth}
  \centering
  \includegraphics[height=1.5in,width=1.7in,angle=0]{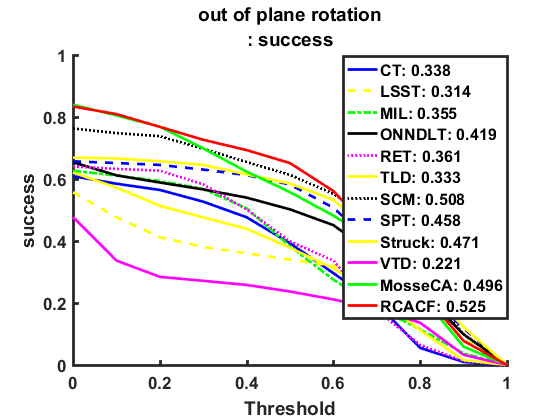}
  \label{fig:sfig2}
\end{subfigure}
\begin{subfigure}{.26\textwidth}
  \centering
  \includegraphics[height=1.5in,width=1.7in,angle=0]{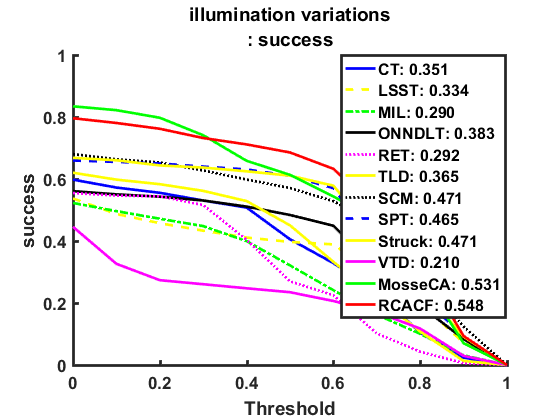}
  \label{fig:sfig2}
\end{subfigure}
\caption{Success plot of various attributes (RCACF - Proposed algorithm (red colour))}
\label{fig:fig}
\end{figure*}

\begin{figure*}
\begin{subfigure}{.26\textwidth}
  \centering
  \includegraphics[height=1.5in,width=1.7in,angle=0]{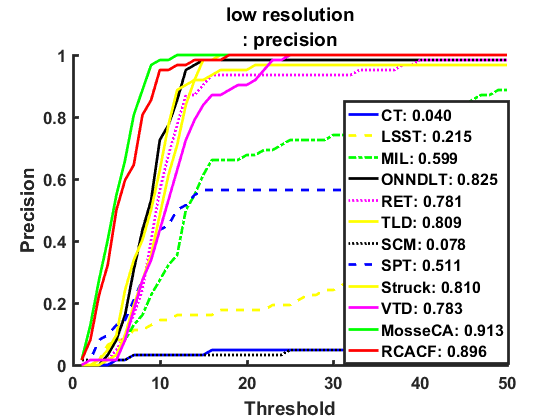}
  \label{fig:sfig1}
\end{subfigure}%
\begin{subfigure}{.2\textwidth}
  \centering
  \includegraphics[height=1.5in,width=1.7in,angle=0]{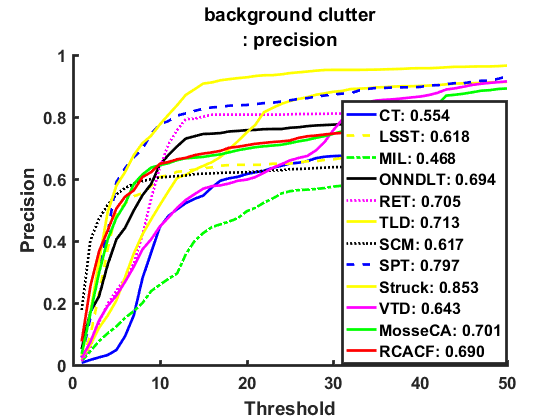}
  \label{fig:sfig2}
\end{subfigure}
\begin{subfigure}{.2\textwidth}
  \centering
  \includegraphics[height=1.5in,width=1.7in,angle=0]{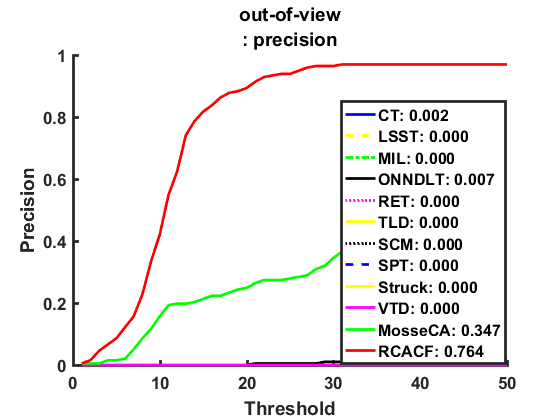}
  \label{fig:sfig2}
\end{subfigure}
\begin{subfigure}{.2\textwidth}
  \centering
  \includegraphics[height=1.5in,width=1.7in,angle=0]{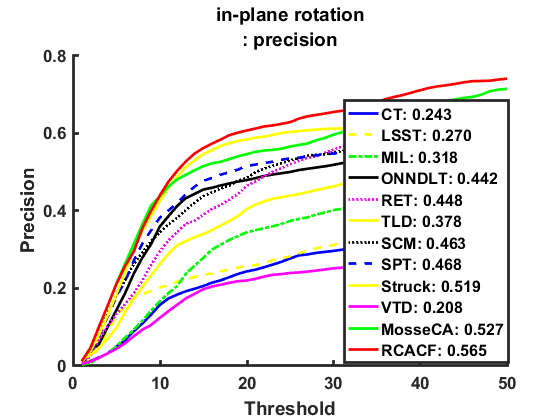}
  \label{fig:sfig2}
\end{subfigure}
\begin{subfigure}{.2\textwidth}
  \centering
  \includegraphics[height=1.5in,width=1.7in,angle=0]{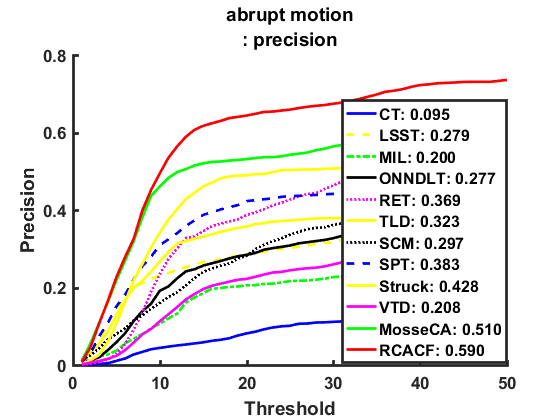}
  \label{fig:sfig2}
\end{subfigure}
\begin{subfigure}{.2\textwidth}
  \centering
  \includegraphics[height=1.5in,width=1.7in,angle=0]{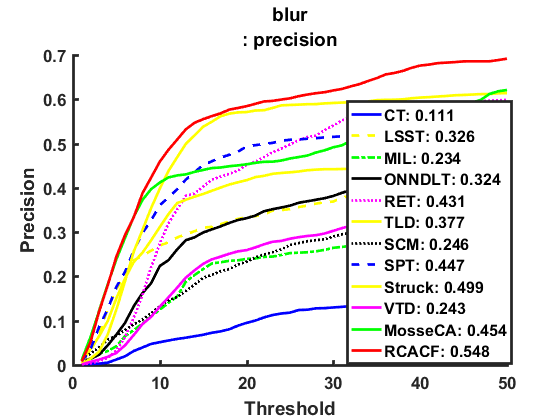}
  \label{fig:sfig2}
\end{subfigure}
\begin{subfigure}{.2\textwidth}
  \centering
  \includegraphics[height=1.5in,width=1.7in,angle=0]{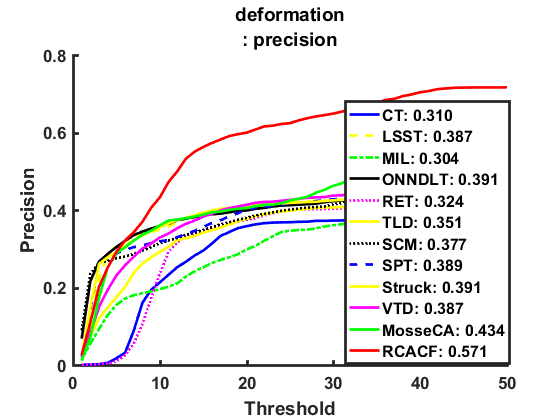}
  \label{fig:sfig2}
\end{subfigure}
\begin{subfigure}{.2\textwidth}
  \centering
  \includegraphics[height=1.5in,width=1.7in,angle=0]{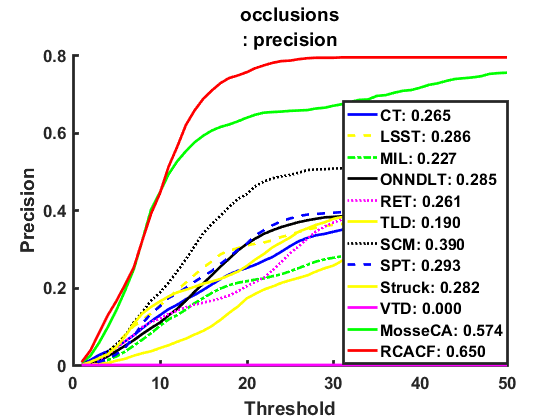}
  \label{fig:sfig2}
\end{subfigure}
\begin{subfigure}{.26\textwidth}
  \centering
  \includegraphics[height=1.5in,width=1.7in,angle=0]{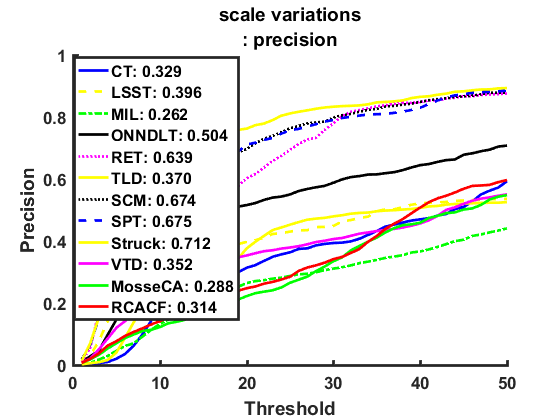}
  \label{fig:sfig2}
\end{subfigure}
\begin{subfigure}{.26\textwidth}
  \centering
  \includegraphics[height=1.5in,width=1.7in,angle=0]{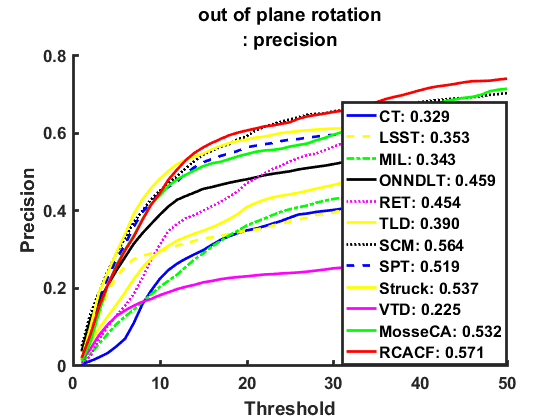}
  \label{fig:sfig2}
\end{subfigure}
\begin{subfigure}{.26\textwidth}
  \centering
  \includegraphics[height=1.5in,width=1.7in,angle=0]{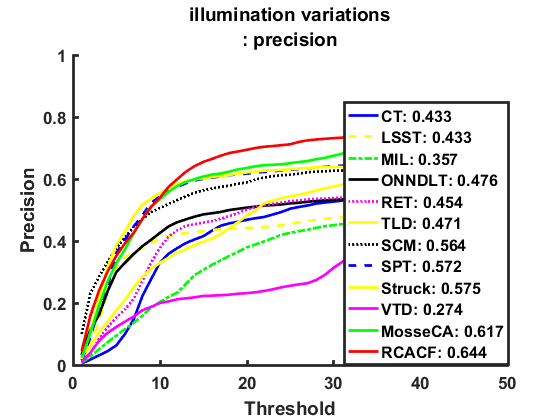}
  \label{fig:sfig2}
\end{subfigure}
\caption{Precision plot of various attributes (RCACF - Proposed algorithm (red colour))}
\label{fig:fig}
\end{figure*}

From Fig 2, it is evident that the proposed approach outperforms all the mentioned state-of-the-art-trackers. The overall precision of trackers such as CT has an accuracy of 36$\%$, LSST has an accuracy of 47$\%$, MIL obtained an accuracy of 36$\%$, ONNDLT has 49.2$\%$, RET has 51.5$\%$, TLD has 48.7$\%$, SCM has 52$\%$, SPT has 60.6$\%$, Struck has 62.3$\%$, VTD has 28.1$\%$, MosseCA has 62.3$\%$ and the proposed algorithm (RCACF) has 64.9$\%$. Similary in case of overall success plot, CT acquired an accuracy of 34.9$\%$, LSST (39.7$\%$), MIL (35$\%$), ONNDLT (42.5$\%$), RET (40.5$\%$), TLD (41.2$\%$), SCM (46.6$\%$), SPT (52.5$\%$), Struck (53.6$\%$), VTD (23.2$\%$), MosseCA (56$\%$) and RCACF (57.8$\%$). Overall performance of the RCACF (Proposed) tracker compared to other trackers shows RCACF (Proposed) provides improved performance compared to all other trackers. 

Fig 3 shows the success plot of various attributes for the considered trackers. For the attribute based analysis of success plot, the proposed algorithm provides an accuracy of 69.8$\%$ for low resolution, 58.6$\%$ for background clutter, 66.4$\%$ for out-of-view, 51.2$\%$ for in-plane-rotation, 53.6$\%$ for abrupt motion, 50.8$\%$ for blur, 52.2$\%$ for deformation, 57.6$\%$ for occlusion, 52.6$\%$ for out-of-plane-rotation, 54.9$\%$ for illumination variation, and 40.1$\%$ for scale variations. Fig 4 shows the attribute based precision plot. The proposed algorithm provides a precision of 89.6$\%$ for low resolution, 69.2$\%$ for background clutter, 76.4$\%$ for out-of-view rotation, 56.6$\%$ for in-plane-rotation, 59$\%$ for abrupt motion, 54.8$\%$ for blur, 57.1$\%$ for deformation, 65$\%$ for occlusion, 31.7$\%$ for scale variations, 57.2$\%$ for out-of-plane rotation and 64.5$\%$ for illumination variation. For most of the attributes the proposed algorithm outperforms the existing trackers. 

\section{Conclusion}
We propose a framework for correlation trackers by incorporating only a single selective background patch and a restoration filter. The algorithm explains the importance of selective background patches and a restored correlation filter in the field of tracking. The qualitative and quantitative analysis shows that the proposed approach performs better tracking compared to other trackers. The algorithm provides an average success score of 86.89$\%$ for 13 data sequences from visual tracking benchmark. The proposed algorithm also shows improved performance for attributes based measures such as motions blur, out of plane rotation, occlusion etc.



\end{document}